%% file: main.tex
\documentclass{article}

     \PassOptionsToPackage{numbers, compress}{natbib}

\usepackage{arxiv}

\usepackage[utf8]{inputenc} 
\usepackage[T1]{fontenc}    
\usepackage[hyperfootnotes=false]{hyperref}       
\usepackage{url}            
\usepackage{booktabs}       
\usepackage{amsfonts, amssymb}       
\usepackage{nicefrac}       
\usepackage{microtype}      
\usepackage[svgnames]{xcolor}         
\usepackage{graphicx}
\usepackage{booktabs}
\usepackage{framed}
\usepackage{caption, subcaption}
\newcommand{\prompt}[2][]{\textbf{#1}\texttt{\textcolor{red}{#2}}}
\newcommand\blfootnote[1]{%
  \begingroup
  \renewcommand\thefootnote{}\footnote{#1}%
  \addtocounter{footnote}{-1}%
  \endgroup
}
\definecolor{shadecolor}{named}{LightGray}

\newcommand{\datasetname}{Magdalena Camera Traps}
\newcommand{\methodname}{WildMatch}

\title{Multimodal Foundation Models for Zero-shot Animal Species Recognition in Camera Trap Images}

\author{%
	Zalan Fabian$^{*\dagger}$\\
 \And
 Zhongqi Miao$^*$\\
 \And
 Chunyuan Li$^\ddagger$\\
 \And
 Yuanhan Zhang$^\circ$ \\
 \And
 Ziwei Liu$^\circ$ \\
 \And
 Andrés Hernández$^{*\blacklozenge}$ \\
 \And
 Andrés Montes-Rojas$^\blacklozenge$ \\
 \And
 Rafael Escucha$^\blacklozenge$ \\
 \And
 Laura Siabatto$^\blacklozenge$ \\
 \And
 Andrés Link$^\blacklozenge$ \\
 \And
 Pablo Arbeláez$^\blacklozenge$ \\
 \And
 Rahul Dodhia$^*$ \\
 \And
 Juan Lavista Ferres$^*$
}

\begin{document}
  \blfootnote{$^*$Microsoft AI for Good Lab, Redmond}
  \blfootnote{$^\dagger$University of Southern California, Los Angeles}
\blfootnote{$^\ddagger$Microsoft Research, Redmond}
\blfootnote{$^\circ$Nanyang Technological University, Singapore}
\blfootnote{$^\blacklozenge$Universidad de Los Andes, Colombia}
\blfootnote{$~~$Corresponding authors: Zalan Fabian ( \texttt{zfabian@usc.edu}) and Zhongqi Miao (\texttt{zhongqimiao@microsoft.com}) }
\maketitle

\input{abstract}
\input{intro}
\input{background}
\input{method}
\input{experiments}
\input{conclusion}

\clearpage
\bibliographystyle{ieee_fullname}
\bibliography{references}

\newpage
\appendix
\input{appendix}
\end{document}

%% file: abstract.tex
\begin{abstract}
Due to deteriorating environmental conditions and increasing human activity, conservation efforts directed towards wildlife is crucial. Motion-activated camera traps constitute an efficient tool for tracking and monitoring wildlife populations across the globe. Supervised learning techniques have been successfully deployed to analyze such imagery, however training such techniques requires annotations from experts. Reducing the reliance on costly labelled data therefore has immense potential in developing large-scale wildlife tracking solutions with markedly less human labor. In this work we propose \methodname{}, a novel zero-shot species classification framework that leverages multimodal foundation models. In particular, we instruction tune vision-language models to generate detailed visual descriptions of camera trap images using similar terminology to experts. Then, we match the generated caption to an external knowledge base of descriptions in order to determine the species in a zero-shot manner. We investigate techniques to build instruction tuning datasets for detailed animal description generation and propose a novel knowledge augmentation technique to enhance caption quality. We demonstrate the performance of \methodname{} on a new camera trap dataset collected in the Magdalena Medio region of Colombia.   
\end{abstract}

%% file: intro.tex
\section{Introduction}
Camera traps are motion-activated remote cameras that are used extensively to monitor wildlife populations. They are deployed worldwide for tasks such as density estimation of animal populations, species inventory and analysis of animal behavior \cite{lucas2015generalised, rahman2017population, caravaggi2017review}. Wildlife monitoring is more important than ever due to the devastating effects of increasing human activity and climate change on natural habitats. Camera traps offer a non-invasive and scalable solution, however the analysis of obtained imagery requires significant effort from wildlife experts \cite{norouzzadeh2018automatically}.

Supervised machine learning approaches have been proposed and successfully deployed for wildlife detection and species classification in camera trap imagery \cite{norouzzadeh2018automatically, tabak2019machine, waldchen2018machine, miao2019insights}. Even though these techniques can help automate much of the visual recognition pipeline, they suffer from the well-known shortcomings of supervised techniques. First, massive amounts of annotated data is required to train the models. As the distribution of camera trap images exhibits strong domain variations (different environment, local species, camera setup), a new dataset needs to be collected for every region. Therefore supervised models are unfit for use-cases where such annotated dataset is not available yet. The problem of data collection is compounded by the fact that expert annotators with specialized knowledge of the local species are needed to label the images. Second, supervised models often lack robustness when deployed on data even slightly different from the training set \cite{zhao2020multi}. Therefore, these models have difficulties when encountering low-quality and corrupted images, both very common in camera trap imagery due to motion blur, low-light conditions and occlusions. 

Multimodal foundation models, such as Multimodal GPT-4 \cite{openai2023gpt}, Flamingo \cite{alayrac2022flamingo}, LLaVA \cite{liu2023visual}, InstructBLIP \cite{instructblip} and Otter \cite{li2023otter} have been the driving force behind the revolution of artificial intelligence recently (see \cite{li2023multimodal} for a comprehensive survey). Large multimodal models (LMMs) ground images to the natural language domain and demonstrate strong capabilities in image understanding and reasoning. Multimodal models such as CLIP \cite{radford2021learning} have proven to have strong zero-shot classification performance and are able to generalize to novel concepts and categories not directly seen in the training set. Moreover, recent work \cite{huang2023language} serves as a strong indicator that in-context language descriptions can guide LMMs to differentiate between fine-grained categories, a crucial requirement in animal species classification. 

\begin{figure}[t]
  \includegraphics[width=\linewidth]{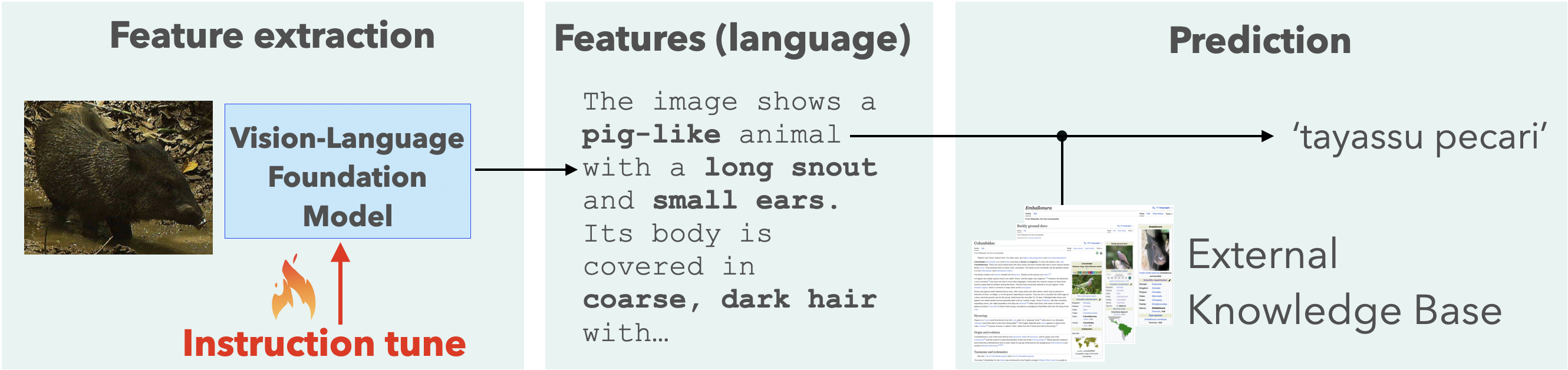}
  \caption{Overview of \methodname{}. We instruction tune a large multimodal model for detailed animal description generation. We extract visual features of camera trap imagery in the form of fine-grained captions. We leverage a knowledge base of animal descriptions to find the best match in order to identify the species. }
  \label{fig:overview}
\end{figure}

In this work we propose \methodname{}, a pipeline for fine-grained zero-shot classification for animal species classification leveraging vision-language foundation models (see overview in Figure \ref{fig:overview}). We extract visual features in the natural language domain from camera trap images in the form of detailed image descriptions. Then, we compare the extracted features to a pre-compiled external knowledge base and output the category with the closest match using a large language model (LLM), a technique we call \textit{description matching}. We observe that out-of-the-box LMMs are incapable of generating detailed enough image descriptions and the captions are often ridden by irrelevant information and hallucinations. To tackle this, we propose an instruction tuning pipeline for detailed animal description generation by injecting common-sense and expert knowledge into vanilla LMM-generated image descriptions. We demonstrate the performance of our pipeline on a new camera trap dataset collected from a novel region in Colombia. \methodname{} shows promising performance without the need for any in-domain training data.

%% file: background.tex
\section{Background and Related Work}
\subsection{Zero-shot learning}
Zero-Shot Learning (ZSL) aims at obviating the need for massive training datasets when encountering new categories by leveraging auxiliary prior knowledge that has been either hand-crafted or potentially learned from seen categories. Conceptually, ZSL methods work by transforming semantic descriptors, such as pre-defined attributes \cite{lampert2009learning, changpinyo2020classifier, atzmon2018probabilistic, palatucci2009zero} or semantic text embeddings \cite{wang2016relational, xian2016latent}, and images into a joint embedding space, where related samples are clustered together. In the context of animal classification, \cite{lampert2009learning} introduces an attribute-based ZSL framework, where visual attributes are manually annotated by human experts, which becomes prohibitive for large-scale datasets.  

\subsection{Text-based knowledge in image recognition}
Natural language descriptions has been successfully incorporated in prior ZSL works. Text-based knowledge has been incorporated in various forms such as visual attributes \cite{huang2021attributes, lampert2009learning, xian2018feature, sariyildiz2019gradient}, WordNet \cite{shen2022k} or Wikipedia articles \cite{paz2020zest, elhoseiny2017link, bujwid2021large}. Human-annotated, short visual descriptions have been successfully used to train multimodal bird \cite{he2017fine, reed2016learning} and flower \cite{reed2016learning} classification models.

Related to our work, \cite{paz2020zest} focuses on improving the quality of textual class descriptions via summarizing the most salient visual features from Wikipedia. However, as opposed to \cite{paz2020zest} that leverages sentence similarities for VRS, we prompt an LLM to summarize the appearance and visual features of species in the articles, lending extra flexibility to our method.  

Another approach is to leverage an LLM to generate visual descriptions \cite{pratt2023does, maniparambil2023enhancing} or descriptors \cite{feng2023leveraging} of categories which is then used as a prompt for multimodal models, such as CLIP. In this work, we also leverage the flexibility of an LLM to generate category descriptions. However, as our goal is fine-grained (genus or species-level) classification of potentially rare species, we opt to obtain the class descriptions by \textit{summarizing} reliable external sources, such as Wikipedia instead of directly relying on the output of a pre-trained LLM that may lack accurate domain-specific knowledge. 

\subsection{Adaptation of foundation models} 
Large language/multimodal foundation models are machine learning models with immense capacity (weights in the order of billions) that have been trained on massive, internet-scale data. Techniques to adapt the rich representations of such models to various domains and downstream tasks has attracted significant attention recently \cite{li2023llava, hu2023rsgpt, sun2023pathasst}. As training foundation models from scratch is prohibitively costly, a flurry of activity has emerged to find efficient ways to adapt pre-trained foundation models. 

\textbf{Parameter-efficient fine-tuning --} Parameter-efficient fine-tuning\cite{hu2021lora, lester2021power} involves only training a small number of extra weights in lieu of updating all parameters.

\textbf{Adapters --} Adapters \cite{yuan2021florence, zhang2023llama} are additional, small models attached to the foundation model that can transform the high-quality representations from the pre-trained model for downstream tasks. 

\textbf{In-context learning -- } In-context learning \cite{brown2020language, radford2019language} refers to adapting a pre-trained model to an unseen task without any training via adding training examples as part of the input. Flamingo \cite{alayrac2022flamingo}, an LMM that uses multimodal in-context learning, can be prompted with interleaved image and text data enabling few-shot adaptation to tasks such as image captioning and visual question-answering. Moreover, it has been observed that LMMs can leverage in-context visual descriptions in fine-grained zero-shot classification. In particular, authors in \cite{huang2023language} add detailed descriptions of bird species to the input prompt allowing the model to correctly differentiate between birds of similar appearance.

\textbf{Instruction-tuning --} Instruction-tuning \cite{ouyang2022training, liu2023visual} allows foundation models to follow natural language instructions and perform various real-world tasks via fine-tuning parts of the pre-trained model on instruction-following data. Instruction-tuning improves the zero-shot capabilities on new tasks and enables the model to provide more accurate and relevant answers. Instruction-tuned LMMs have demonstrated success in multimodal conversations, image understanding and visual reasoning \cite{liu2023visual, li2023otter, instructblip}.

\subsection{Challenges of camera trap imagery}
Analyzing camera trap images often poses a formidable challenge even for human experts \cite{norouzzadeh2018automatically}. First, as opposed to generic images of animals on the internet or in benchmark classification datasets where the animal is the focus of the photo, in camera trap images it is very common to have an animal that is only partially visible, far away, or very close up to the camera. This results in tremendous loss of visual information that requires models with a robust and detailed understanding of the animal's appearance for successful recognition. Second, as some animals are only active at night, it is common to have only low-light, motion blurred and noisy images of some species. Even though supervised models can overfit to specific artifacts in such images (for instance the motion blur of flying bats), relying on such spurious correlations for recognition undermines the trust of practitioners and may lead to unexpected errors. Due to the distribution discrepancy between the generic animal images on the internet and the more obscure wildlife camera trap imagery, foundation models pre-trained on the former need to be adapted for downstream image recognition tasks on the latter.

%% file: method.tex
\section{Method}
In this work we propose \methodname{}, a zero-shot animal species classification pipeline that leverages natural language descriptions rather than vision representations directly. The key component of traditional supervised vision models is a feature extractor that learns the relevant visual features in images from a training set. The extracted vision feature vectors are then passed through a classifier head that maps the vision features to discrete class labels. In stark contrast, our extracted features are \textit{in natural language} in the form of image captions obtained from a LMM. The extracted natural language description is then compared with descriptions in an external knowledge base (summarized from online sources) and the entry with the closest match is selected as our final prediction. We introduce our matching technique in Section \ref{sec:match}. The efficiency of the proposed pipeline therefore depends on the two key components of our framework: 1) the quality of the knowledge base and 2) the quality of captions obtained from the LMM.

\subsection{Building a knowledge base}
We obtain a list of all species appearing in the publicly available LILA BC Camera Traps \cite{LILA}, a collection of $18$ camera trap datasets with species annotations. We obtain textual descriptions of each species by scraping their corresponding Wikipedia article. As the articles contain a lot of information irrelevant for visual recognition, we perform \textit{visually relevant extractive summarization} (VRS) \cite{paz2020zest}, the task of extracting sentences with visually relevant information. Details on prompts for VRS and text post-processing can be found in Appendix \ref{apx:wiki_sum}.

\subsection{Shortcomings of out-of-the-box LMMs}
\begin{figure}[t]
  \includegraphics[width=\linewidth]{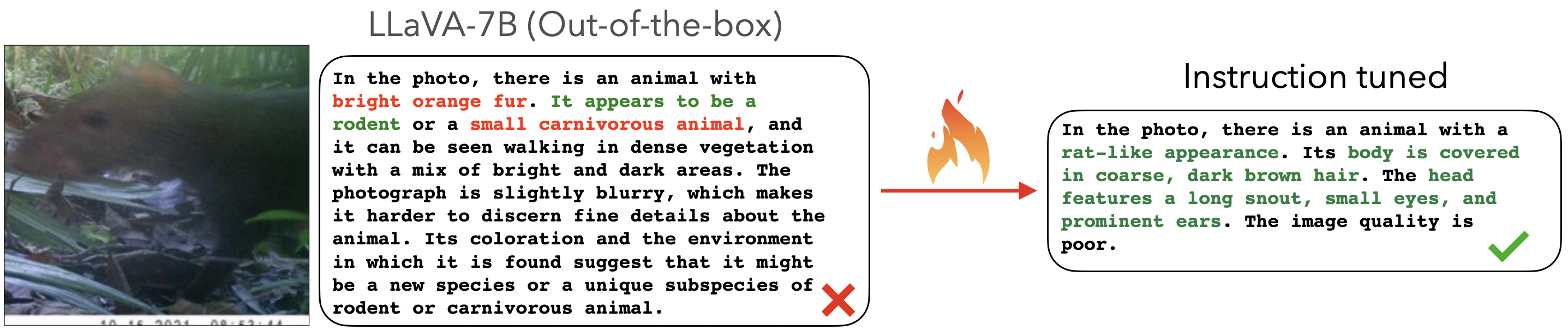}
  \caption{The general-domain LMMs trained on web data  are unable to generate captions suitable for species identification out-of-the-box: the captions often miss relevant details, add excessive irrelevant information and often contain hallucinations. Domain-specific instruction tuned LMMs generate more relevant and succinct captions. }
  \label{fig:LMM_cap}
\end{figure}

The performance of our pipeline strongly depends on the quality of image captions obtained from the LMM. However, we observe that most currently available LMMs (e.g. LLaVA, InstructBLIP or Otter) out-of-the-box are not suitable for extracting relevant details for species identification (Figure \ref{fig:LMM_cap}, left). First, LMM captions often miss relevant and characteristic features of animals that are necessary for correctly identifying the species. Second, out-of-the-box LMMs tend to generate excessive irrelevant information (e.g. speculations  about the image or comments on the beauty of the scene) that has no use to or even mislead the matching algorithm. Lastly, hallucinations are very common in LMM captions \cite{li2023evaluating}, such as hallucinated colors, body parts, other animals or entities appearing in the image. 

LMMs have been pretrained on massive amounts of multimodal data scraped from the internet, and as such they are imbued with a rich understanding of visual concepts. However, as these models have been mostly trained on generic 'internet data', they are not suitable for the specific task of generating detailed descriptions of animals with the focus on species identification. Therefore, we propose adapting LMMs for detailed animal description generation via instruction tuning. Our goal with instruction tuning is to guide the model to generate captions that leverage language used by experts in describing animals. 

\subsection{Instruction tuning for detailed visual descriptions}
In order to instruction tune the model, we need paired data of wildlife imagery with detailed enough captions for species identification. Even though some datasets exist with species annotations (ImageNet \cite{deng2009imagenet}, iNaturalist \cite{van2018inaturalist}), bounding boxes (most LILA Camera Traps \cite{LILA}) or even short captions (Caltech-UCSD Birds \cite{wah2011caltech}), there is no available large-scale dataset with detailed visual descriptions of animals. In fact, collecting such a dataset requires significant effort from experts far exceeding simple species annotations. We tackle the challenge of instruction tuning data collection from two directions.

First, we collect a small dataset of human-captioned wildlife imagery. In order to circumvent the need for biologists for annotations, we extract visual features of species from their corresponding Wikipedia articles and ask human annotators to select the features that are visible in the image. We collect approximately $1.5k$ manually annotated samples from volunteers. Human-annotated data collection is detailed in Appendix \ref{apx:manual}.

Second, we propose a scalable and automatic method of generating captions for species identification via augmenting vanilla out-of-the-box LMM captions with common-sense and expert knowledge. In particular, we process the vanilla LMM captions the following ways:
\begin{itemize}
    \item \textbf{Color filtering --} LMMs often hallucinate colors even on grayscale or night time imagery that may entirely derail species identification. To tackle this, we filter out any color related information if we detect that it should not be identifiable from the image. 
    \item \textbf{Expert knowledge --} We inject information from the species' Wikipedia article pertaining to visual characteristics of the animal that might be visible in the image. Furthermore, we remove details that directly contradict the knowledge base. This step enhances the captions with relevant expert terminology used to describe animal species.
\end{itemize}
We perform these steps using LLM prompting (details in Appendix \ref{apx:know_aug}). We refer to the obtained descriptions as \textit{pseudo-captions} (analogous to pseudo-labels in semi-supervised learning). In order to build the instruction tuning dataset, we generate single-turn conversations from the animal descriptions, where the instruction is sampled from various prompts asking to describe visual characteristics of the animal in the photo (details in Appendix \ref{apx:it}).
\begin{figure}[t]
  \includegraphics[width=\linewidth]{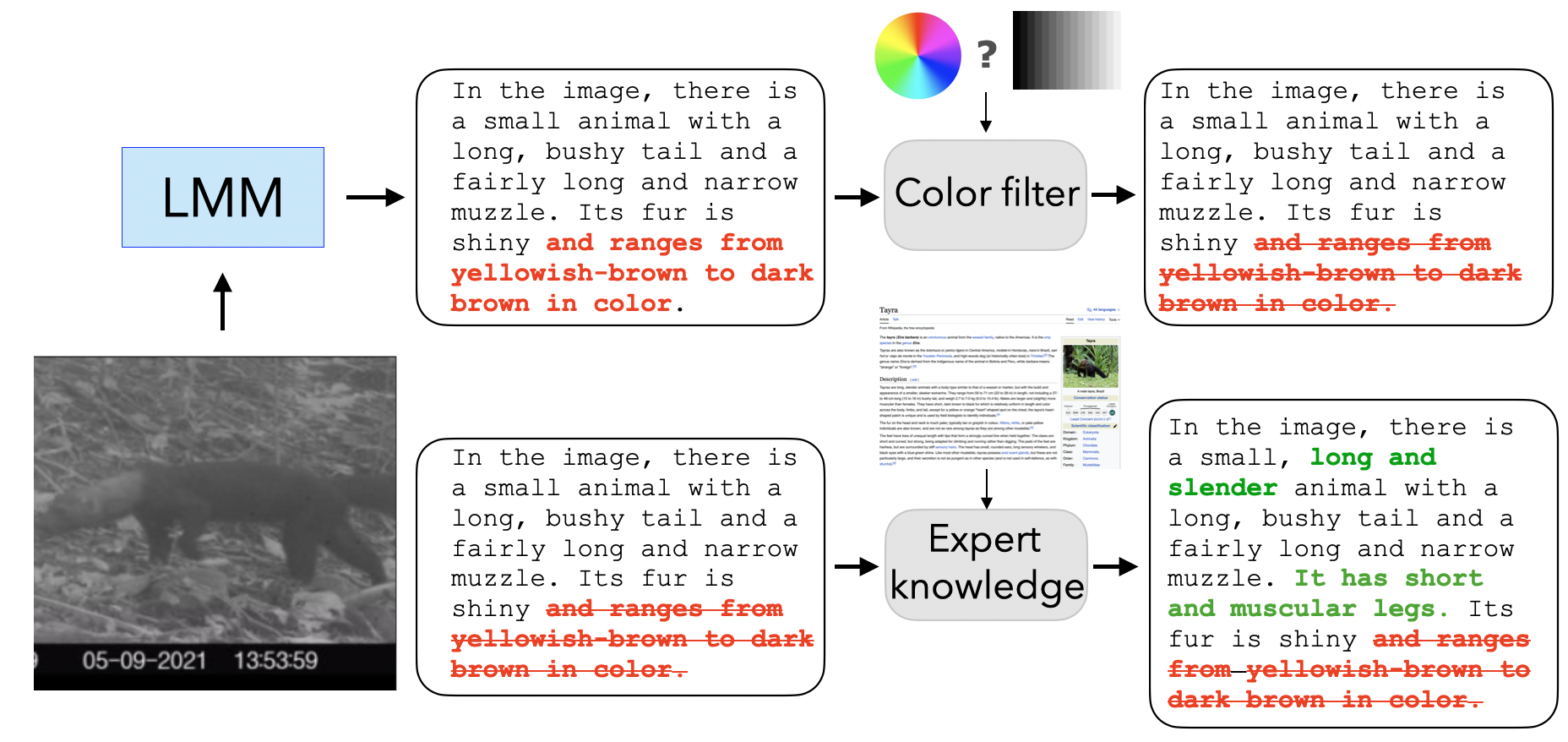}
  \caption{To improve caption quality of LMMs on wildlife images, we perform two processing steps on the vanilla LMM captions: 1) we filter out color related information if we detect that colors are not discernible in the image and 2) we inject expert knowledge from the species' Wikipedia article pertaining to visual characteristics of the animal that might be visible in the image. We leverage the resulting augmented captions to instruction tune a LMM for detailed animal description generation.}
  \label{fig:cap_aug}
\end{figure}

We apply an iterative instruction tuning scheme, where we train multiple generations of models. The first generation is instruction tuned on pseudo-captions generated by out-of-the-box LMMs, or the human-annotated samples. Subsequent generations are then instruction tuned on pseudo-captions from the previous generation's LMM captions.

\subsection{Animal species classification via description matching \label{sec:match}}
The key idea of our proposed pipeline is to caption camera trap images of animals via LMMs and subsequently find the closest matching species in a knowledge base. In particular, we sample a detailed description from our instruction tuned LMM and compare the generated caption to the description of each species in the knowledge base via a large language model (in this work GPT-4). The LLM is instructed to answer with the name of the species which best matches the LMM description. As we provide the knowledge base to the LLM before eliciting an answer, the matching algorithm can be thought of as an example of in-context learning. Details of the prompt used to instruct the LLM to perform the matching can be found in Appendix \ref{apx:match}. 

We further improve the performance of our method through a technique analogous to self-consistency in chain-of-thought reasoning \cite{wang2022self}. Self-consistency appeals to the intuition that the correct answer can be obtained from multiple valid reasoning paths. In particular, we sample $N$ independent captions from the instruction tuned LMM and match each descriptions individually. We obtain the final prediction via majority voting between the $N$ resulting predictions.

\subsection{Hierarchical prediction scheme \label{sec:hierarch}} 
\begin{figure}[t]
\centering
  \includegraphics[width=0.7\linewidth]{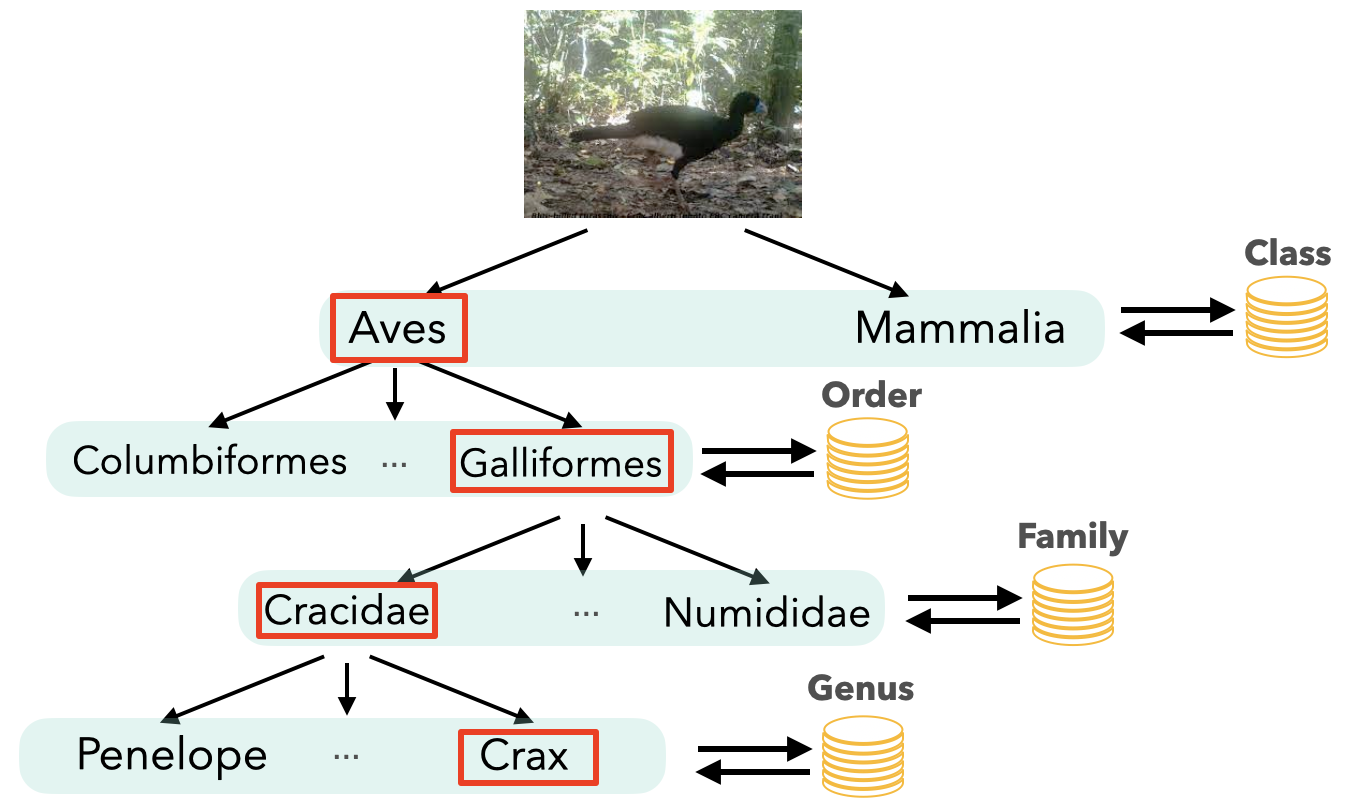}
  \caption{We predict the fine-grained class (species or genus) via hierarchical predictions traversing the taxonomy tree.}
  \label{fig:hierarch_pred}
\end{figure}

As the LLM used in description matching takes the complete knowledge base describing each of the categories as input, matching becomes infeasible with large knowledge bases that exceed the token limitations of the LLM. As animal taxonomy is inherently hierarchical, species recognition lends itself to hierarchical classification. Thus, we propose breaking down the fine-grained prediction task (species or genus) with a large number of potential labels into a sequence of smaller hierarchical predictions each with a manageable knowledge base (Figure \ref{fig:hierarch_pred}). In particular, we build separate knowledge bases for each taxonomic rank (class, order, family, genus, species) and perform description matching top-down, only including categories that fall under the taxonomic group predicted in the previous step.

Performing description matching to directly predict the most fine-grained category (genus or species) becomes prohibitive with a large number of knowledge base entries (number of potential output labels), as the LLM input includes the complete knowledge base. 

Therefore, instead of predicting the most fine-grained label directly (species or genus), we use hierarchical predictions to iteratively narrow down the possible fine-grained classes. As animal taxonomy is inherently hierarchical, species recognition lends itself to hierarchical classification. One may build separate knowledge bases for each taxonomic rank (class, order, family, genus, species) and perform the matching technique on a reduced set of categories. We terminate the iterative prediction scheme when the potential number of fine-grained categories have been narrowed down to a pre-determined (small) number that can be directly handled by the LLM.  

\subsection{Confidence of model predictions}
Typical supervised classifiers in computer vision, such as ResNet and others, output a probability distribution over the label space. Subsequently, the label with the highest assigned probability is used as the final model prediction. This probability is often interpreted as the model's confidence in its prediction. Quantifying model confidence is an essential tool in evaluating model calibration and assessing the reliability of classification results. A well-known shortcoming of supervised classifiers is their poor calibration, as they are often overly confident in their predictions, undermining the trust of practitioners in such models.

As our proposed technique does not explicitly assign a probability distribution to labels, we leverage the self-consistency framework introduced in Section \ref{sec:match} to obtain an approximation. In particular, we sample $N$ predictions, and use the frequency of predicted labels as a proxy for the output probability distribution. In other words, assuming an  input image $x$ and a finite set of discrete labels $\mathcal{Y}$, the probability assigned to label $i \in \mathcal{Y}$ is $\frac{n_i(x)}{N}$, where $n_i(x)$ is the count of predictions with label $i$ out of $N$ with input image $x$. Then, our final prediction is $\hat{y}(x) = arg\max_{i\in\mathcal{Y}} \frac{n_i(x)}{N}$ and the assigned prediction confidence is $c(x) = \frac{n_{\hat{y}}(x)}{N}$. Using this simple proxy for confidence, we observe that our proposed method is better calibrated than supervised models (more details in Section \ref{sec:exp}).

\subsection{Human-in-the-loop classification}
Samples that are challenging for a classifier commonly occur in camera trap datasets due to heavy image corruptions (motion blur, low resolution crops) and partial visibility of the animals (occlusions, out-of-frame body parts). Supervised classifiers are prone to overestimating their own performance, that is they tend to assign high confidence even to wrong predictions hindering the model's ability to anticipate errors.

\begin{figure}
\centering
\begin{subfigure}[t]{.65\textwidth}
  \centering
  \includegraphics[width=0.9\linewidth]{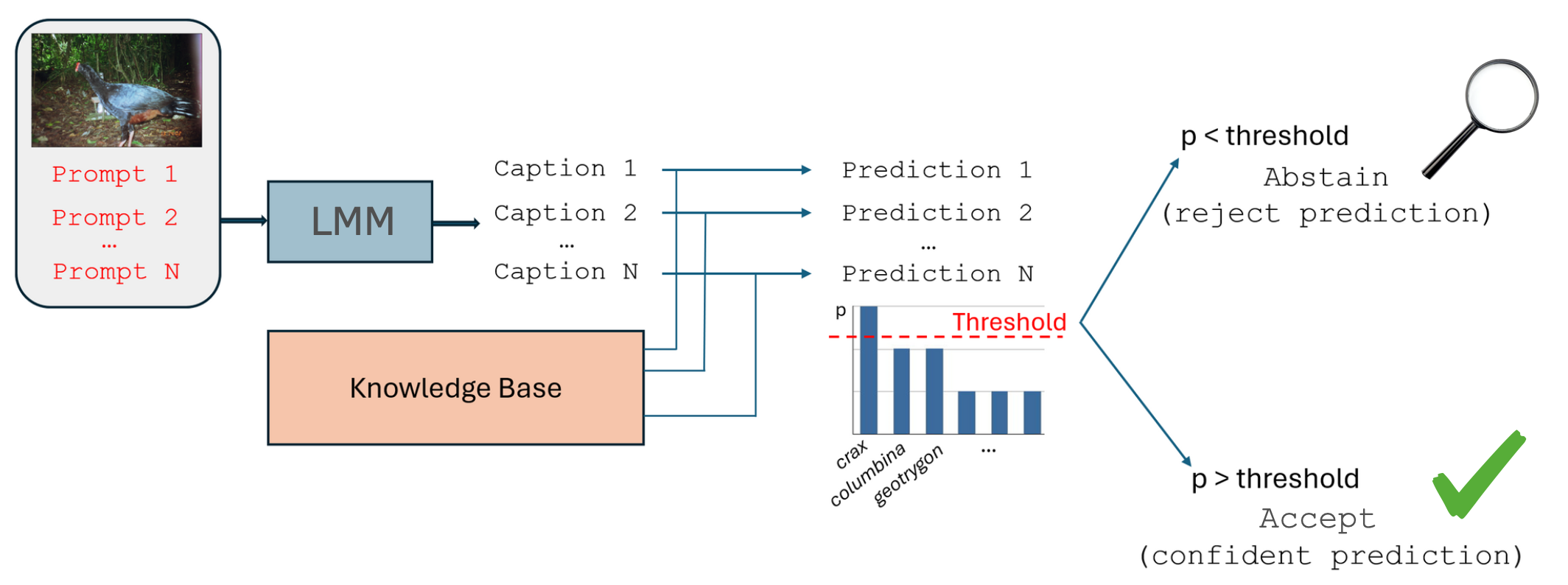}
  \caption{We adjust a confidence threshold to determine whether a given prediction is to be accepted or the model should abstain from prediction and fall back to human labelling.\label{fig:acc_rej}}
\end{subfigure}\hfill
\begin{subfigure}[t]{.3\textwidth}
  \centering
  \includegraphics[width=0.99\linewidth]{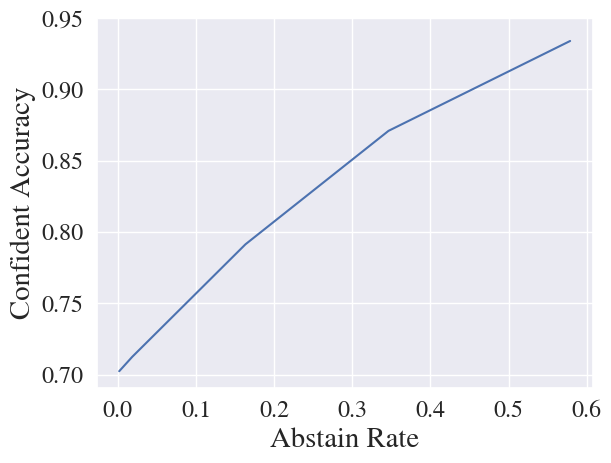}
  \caption{Confident accuracy vs. abstain rate. on the \datasetname{} dataset. \label{fig:acc_rej_plot}}
\end{subfigure}
\caption{We assign confidence to model predictions based on empirical label counts. Then, we filter predictions and route samples with low confidence predictions to experts for further evaluation.}
\end{figure}

As we empirically observe that \methodname{} is better calibrated than supervised models, we leverage the prediction confidence to detect hard samples (Figure \ref{fig:acc_rej}). In particular, for a given threshold $p$ we categorize a prediction high-confidence if the assigned confidence is higher than $p$, otherwise we consider it low-confidence. We propose a zero-shot human-in-the-loop classification framework, in which high-confidence predictions are accepted and the model abstains from prediction on low-confidence samples, routing them to an expert for further evaluation. 

The threshold $p$ serves as a flexible knob to trade off human effort for increased accuracy. Let $\mathcal{D}_{test}$ denote the test dataset with $(x_i, y_i) \in \mathcal{D}_{test}$, where $y_i$ denotes the true label of image $x_i$. Furthermore, let $\mathcal{A} = \{x_i \in \mathcal{D}_{test}| c(x_i) \geq p\}$ the set of accepted (high-confidence) samples and $\mathcal{A}_{corr} = \{x_i \in \mathcal{A}| \hat{y}(x_i) = y_i\}$ the set of accepted samples with correct prediction. We define the \textit{abstain rate} (AR) as $\frac{|\mathcal{D}_{test}| - |\mathcal{A}|}{|\mathcal{D}_{test}|}$ and the \textit{confident accuracy} (CA) as $\frac{|\mathcal{A}_{corr}|}{|\mathcal{A}|}$. Clearly, as we increase $p$, the model abstains from prediction on more and more samples that are in turn routed to an expert. At the same time, the model can focus on samples on which it has higher confidence, leading to increased CA.  

\subsection{Sequence-level predictions \label{sec:seq}}
Camera traps are activated by motion in the field of view of the camera. An animal passing by may trigger the camera multiple times, thus it is typical to capture several frames of the same animal. Akin to a form of natural data augmentation, these frames provide different views of the same animal, which one may leverage to further improve species recognition. 

Specifically, we create sequences by grouping frames that have been captured close in time. Then, we perform description matching with self-consistency on each frame in the sequence individually and aggregate the results to provide a single prediction for the whole sequence. In particular, for $F$ frames in the sequence and $N$ samples per frame due to self-consistency, we perform description matching  $F \cdot N$ times and obtain the final prediction via majority voting. This technique greatly improves classification performance as each view provides additional visual cues for identifying the species.  

%% file: experiments.tex
\section{Experiments\label{sec:exp}}
\textbf{Model --} We leverage the LLaVA framework for our LMM and instruction tune LLaVA-7B models on our animal description instruction tuning datasets.

\textbf{Instruction tuning data --} We create two splits from LILA BC Camera Traps datasets: 1) \textit{LILA-10k} consists of image crops of the approx. $10k$ highest confidence bounding boxes from MegaDetector and all human-annotated bounding boxes where we limit the maximum number of crops per species to $25$; 2) \textit{LILA-40k} is created similarly but with maximum number of crops per species set to $100$. Moreover, we create a human-annotated dataset of approx. $1.5k$ images, where we sample image crops of the $2$ highest confidence bounding boxes for each species appearing in the LILA BC datasets, and also add the Wikipedia image from their corresponding article. We refer to this dataset as \textit{Manual-1.5k}.

\textbf{Evaluation data --} We evaluate our technique's performance on a new camera trap dataset collected in the Magdalena Medio region in Colombia. As the dataset has not yet been publicly released, the LMM has not seen our validation dataset neither during pretraining nor during instruction tuning. 

The dataset consists of 41904 samples (33569 train, 8335 validation) with 36 discrete labels with genus level annotation. The image resolution is $256\times256$.
For benchmarking, we sample 20 classes with imbalanced data distribution to simulate a realistic camera trap setting. We refer to this dataset as \textit{\datasetname{}}. We use micro and macro classification accuracy as evaluation metrics, where micro accuracy refers to accuracy in the traditional sense and macro accuracy is the average of individual class accuracies across all classes.

\textbf{Instruction tuned models --} We investigate the value of both the number of samples used in instruction tuning as well as the quality of descriptions. To this end, we instruction tune various models on both pseudo-captioned and human-annotated data (see Table \ref{tab:models}). We build a knowledge base of all species occuring in the LILA datasets for expert knowledge augmentation. We use $4 \times$ A100 GPUs for instruction tuning. 

\begin{table}
\parbox{.45\linewidth}{
\centering
\resizebox{7.5cm}{!}{
   \begin{tabular}{lcc}
    \hline
    \textbf{Model}           &\textbf{Pseudo-captioning LMM} & \textbf{Instruction tuning data} \\ \hline
    \textit{LLaVA-7B-M-gen1} & {\color[HTML]{0000FF} -}                            & {\color[HTML]{000000} Manual-1.5k}                    \\ 
    \textit{LLaVA-7B-M-gen2} & {\color[HTML]{000000} LLaVA-7B-M-gen1}              & {\color[HTML]{000000} LILA-40k}                       \\ 
    \textit{LLaVA-7B-P-gen1} & LLaVA-7B                                            & LILA-10k                                              \\ 
    \textit{LLaVA-7B-P-gen2} & LLaVA-7B-P-gen1                                     & LILA-40k                                            \\   \bottomrule \vspace{0.025cm}
    \end{tabular}
    }
\caption{Summary of models we instruction tune for animal description generation.\label{tab:models}}
}
\hfill
\parbox{.5\linewidth}{
\centering
\resizebox{5.5cm}{!}{
		\begin{tabular}{ lcc }
			\toprule
			&\multicolumn{2}{c}{\textbf{\datasetname{}}} \\
			\cmidrule{2-3}
			\textbf{Method} &Micro acc.&Macro acc. \\
			\midrule
			Supervised & 93.94\%&85.50\% \\
			\hline
            CLIP (category name)  & 35.85 \%& 33.13 \% \\
            CLIP (description)  &44.28 \%&36.16 \% \\
            LLaVA-7B & 54.96\%& 50.76\%\\
			LLaVA-7B-M-gen1 &62.00\%&59.28\% \\
			LLaVA-7B-M-gen2 &\textbf{70.12\%}&\textbf{64.75\%} \\
   			LLaVA-7B-P-gen1 &65.23\%&61.57\% \\
			LLaVA-7B-P-gen2 &\underline{69.94\%}&\underline{64.36\%} \\
			\bottomrule \vspace{0.025cm}
		\end{tabular}
    }
\caption{Results on the \datasetname{} dataset.\label{tab:results}}
}
\end{table}

\textbf{Inference setting --} We build separate knowledge bases for the \datasetname{} dataset, consisting of summaries of each genus appearing in the dataset. We leverage GPT-4 for description matching and we utilize self-consistency (SC) with majority voting to obtain the final predictions. We use $N=5$ independent samples in the experiments unless mentioned otherwise.

\textbf{Main results --} An overview of our main results is depicted in Table \ref{tab:results}. We train a ResNet-50 classifier on the full dataset to provide a supervised baseline. For a zero-shot baseline, we use CLIP (ViT L/14) where the text embeddings are either category names (genus in this case) or the description of the category from the knowledge base. The latter is closely related to CuPL\cite{pratt2023does}, however the class descriptions are summarized from Wikipedia, and not directly obtained from an LLM. We observe significant improvement over the naive CLIP baseline of close to $+25\%$ in a 20-way classification problem using our best instruction tuned model. Furthermore, generation 2 models consistently outperform their predecessors indicating the utility of additional instruction tuning data and improved data quality. However, we observe only a small gap between generation 2 models that have been trained on either pseudo-captions or human annotations in the first generation. 

\textbf{Impact of instruction tuning data quality --} As collecting human annotated wildlife imagery is costly, it is crucial to investigate how our automatic captioning technique with knowledge augmentation compares to human annotated captions. To this end, we create train and test splits from \textit{Manual-1.5k} and instruction tune two models: one on human-annotated captions of the train split and a second on pseudo-captions of the same split. Pseudo-captions are generated via out-of-the-box LLaVA-7B and our knowledge augmentation method. We evaluate the generated captions quantitatively via classification accuracies when used in our pipeline, and qualitatively via GPT scoring. In particular, we prompt GPT-4 to score the generated caption based on how close they are to the corresponding human annotated ground truth caption. We define two scores: 1) relevance score, which measures how much of the information in the ground truth caption is captured in the generated caption and 2) hallucination score, which assigns higher scores to captions that have less excess details compared to the ground truth. More details and the specific prompts are in Appendix \ref{apx:scores}. We evaluate these qualitative scores on the test split of \textit{Manual-1.5k}. The results are summarized in Table \ref{tab:quality}. Overall, we observe consistent improvement in all metrics when using human-annotated data compared to automatically generated pseudo-captions for instruction tuning. Our experiments indicate that collecting large-scale human captioned wildlife image datasets may enable further improvements in the zero-shot performance of our proposed pipeline.

\begin{table}
\parbox{.58\linewidth}{
\centering
	\resizebox{9.5cm}{!}{
		\begin{tabular}{ lccccc }
			\toprule
			&\multicolumn{2}{c}{\textbf{\datasetname{}}}& & \multicolumn{2}{c}{\textbf{Qualitative}} \\
			\cmidrule{2-3}\cmidrule{5-6}
			\textbf{Instruction tuning data} &Micro acc.&Macro acc.& & Relevance ($\uparrow$) & Hallucination ($\uparrow$) \\
			\midrule
            None  & 54.96 \%&50.76 \% & & 3.98 (1.92) & 4.30 (1.83) \\
            Pseudo-captions  & 57.20 \%&57.10 \% & & 4.25 (1.89) & 4.41 (1.77) \\
			Manual captions &\textbf{62.00\%}&\textbf{59.28\%} & & \textbf{5.10} (1.98) & \textbf{5.33} (1.85) \\
			\bottomrule
		\end{tabular}
	}
    \vspace{1mm}
 \caption{Experiments on the effect of instruction tuning data quality on performance. We evaluate relevance and hallucination scores via GPT-4 prompting (higher the better). Standard deviation of scores in parentheses.\label{tab:quality}}
}
\hfill
\parbox{.38\linewidth}{
\centering
	\resizebox{5.5cm}{!}{
		\begin{tabular}{ lccc }
			\toprule
			&\multicolumn{3}{c}{\textbf{\datasetname{}}}\\
			\cmidrule{2-4}
			\textbf{Method} &ECE&MCE&ACE \\
			\midrule
            Supervised  & \textbf{0.0316}&0.3920 & 0.1490 \\
            \methodname{}(ours)  & 0.0406&\textbf{0.0872} & \textbf{0.0118} \\
			\bottomrule
		\end{tabular}
	}
    \vspace{1mm}
 \caption{Calibration of our proposed method compared with a strong supervised baseline (ResNet50). ECE: Expected Calibration Error, MCE: Maximum Calibration Error, ACE: Average Calibration Error. \label{tab:calib}}
}
\end{table}

\begin{figure}
\centering
\begin{subfigure}{.5\textwidth}
  \centering
  \includegraphics[width=0.8\linewidth]{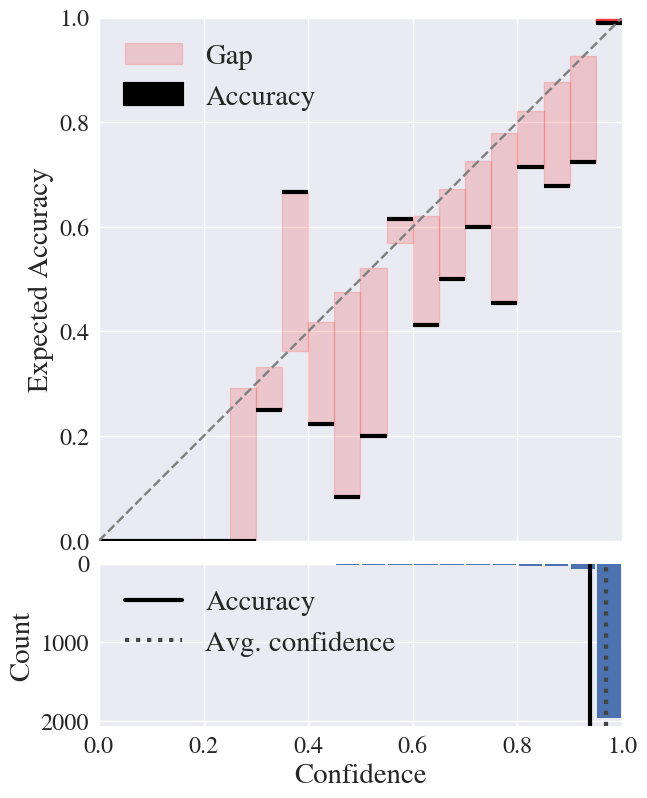}
  \caption{Supervised\label{fig:calib_sup}}
  \label{fig:sub1}
\end{subfigure}%
\begin{subfigure}{.5\textwidth}
  \centering
  \includegraphics[width=0.8\linewidth]{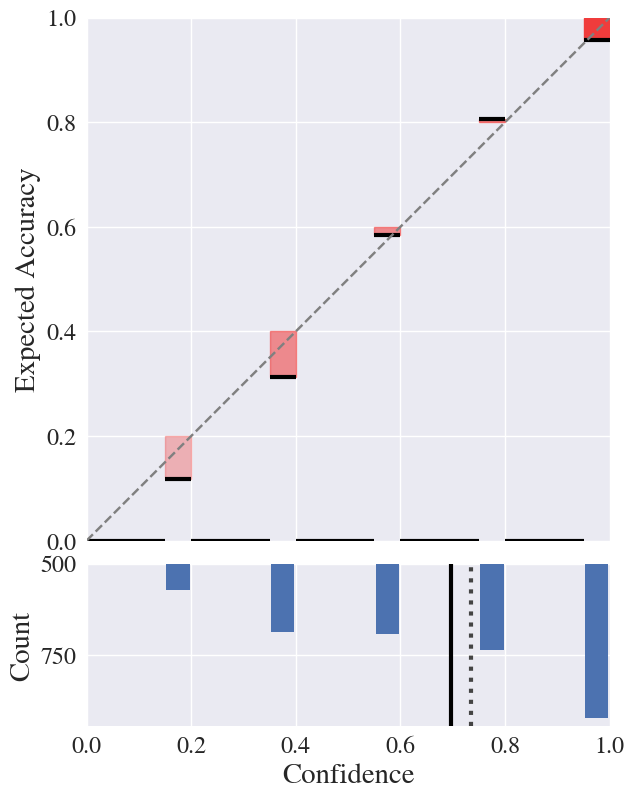}
  \caption{Zero-shot (ours)\label{fig:calib_ours}}
  \label{fig:sub2}
\end{subfigure}
\caption{Calibration comparison via reliability diagrams over 20 confidence bins. Supervised models tend to have poor calibration with over-confident predictions. Our proposed zero-shot method is better calibrated than the supervised baseline in terms of multiple calibration metrics (see Table \ref{tab:calib}).\label{fig:calib}}
\label{fig:test}
\end{figure}

\textbf{Model calibration --} We compare the calibration of our proposed zero-shot framework to a supervised baseline (ResNet-50). Supervised models tend to be overly confident in their predictions, even when they misclassify a sample.  This property is undesired and undermines the trust of experts in the model, regardless of their nominal performance. Model calibration is often evaluated via reliability diagrams that depict the confidence of predictions against their accuracy over discrete bins. We observe that the supervised model is poorly calibrated with over-confident predictions on the \datasetname{} dataset (Figure \ref{fig:calib_sup}). On the other hand, \methodname{} can more accurately assess the reliability of its predictions (Figure \ref{fig:calib_ours}).

We evaluate model calibration in terms of $3$ popular metrics \cite{guo2017calibration, neumann2018relaxed}: Expected Calibration Error (ECE), Maximum Calibration Error (MCE) and Average Calibration Error (ACE) (Table \ref{tab:calib}). See Appendix \ref{apx:calib} for further details of calibration metrics. \methodname{} markedly outperforms the supervised model in terms of MCE and ACE. As opposed to the ECE metric, MCE and ACE quantifies calibration error irrespective of bin counts and thus are better suited for safety-critical and real-world applications where worst-case performance is critical.  

\textbf{Human-in-the-loop classification --} We investigate the performance gain achieved by detecting and routing difficult examples to experts. In particular, we vary the confidence threshold $p$ that determines whether the prediction should be accepted or the model should abstain from prediction and route the sample to a human. We plot AR as a function of CA in Figure \ref{fig:acc_rej_plot}. We observe a steady increase in CA as $p$ (and thus AR) increases, with above $85 \%$ classification accuracy on $70\%$ of the dataset without any training data.

\textbf{Sequence-level predictions --} As described in Section \ref{sec:seq}, camera traps often capture a sequence of frames as the animal passes in front of the camera. We leverage this information to further boost classification accuracy. More specifically, given $F$ number of frames in a sequence, we sample $N$ predictions for each frame via self-consistency, resulting in $F \cdot N$ total number of predictions per sequence. We perform perform majority voting across all $F \cdot N$ predictions and obtain a sequence-level predicted label $\hat{y}_{seq}$. Then, we assign $\hat{y}_{seq}$ as the predicted label to each frame within the sequence. Using this simple scheme, we improve the micro accuracy from $70.12\%$ to $77.54\%$ and the macro accuracy from $64.75\%$ to $71.73\%$ on the \datasetname{} dataset using our best performing model (LLaVA-7B-M-gen2). This constitutes a relative performance improvement of more than $+10\%$ in each metric without additional compute cost. Therefore, we conclude that leveraging sequence information provides significant improvement to zero-shot classification performance and fits naturally into our proposed framework. 

%% file: conclusion.tex
\section{Conclusion}
In this work we have introduced \methodname{}, a novel zero-shot classification framework for wildlife species recognition leveraging multimodal foundation models. We instruction tune vision-language models for detailed animal description generation and utilize a large language model to match the generated description to a pre-compiled external knowledge base. As datasets with detailed enough descriptions of animals are lacking, we propose a novel pipeline for leveraging a combination of human-annotated data and automatically generated (knowledge augmented) captions for instruction tuning. We evaluate the quality of generated descriptions both via their utility in our zero-shot classification framework, and via GPT scoring methods. Our instruction tuned models far surpass the performance of naive zero-shot baselines and out-of-the-box models in animal species classification. The main limitation of the current framework is its high compute requirement compared to supervised models, which we address in future work.

%% file: appendix.tex
\section{Wikipedia article summarization \label{apx:wiki_sum}}
We build the knowledge base from Wikipedia articles of species that are present in either of the LILA BC Camera Trap datasets or the Magdalena dataset used for evaluation. We extract the page summary and any sections where the section title contains any of the following words: \textit{description, characteristics, appearance, anatomy}. We use GPT-4 to summarize features of the animal that may be visible in a photograph, but without referring to specific physical measurements of the animal as those are typically not possible to determine from an image (exact height, weight etc). The following prompt is used for the summarization:
\begin{shaded}
    \noindent \prompt[System message: ]{You are an AI assistant specialized in biology and providing accurate and detailed descriptions of animal species.}
    
    \prompt[Prompt: ]{You are given the description of an animal species. Provide a very detailed description of the appearance of the species and describe each body part of the animal in detail. Only include details that can be directly visible in a photograph of the animal. Only include information related to the appearance of the animal and nothing else. Make sure to only include information that is present in the species description and is certainly true for the given species. Do not include any information related to the sound or smell of the animal. Do not include any numerical information related to measurements in the text in units: m, cm, in, inches, ft, feet, km/h, kg, lb, lbs. Remove any special characters such as unicode tags from the text. Return the answer as a single paragraph. Species description: <WIKI\_ARTICLE> Answer:}
\end{shaded}
\section{Knowledge augmentation \label{apx:know_aug}}
We use GPT-4 to enhance the quality of LMM-generated captions with external knowledge and common sense. 

\textbf{Color filtering --} We detect low color variation in images via the condition
\begin{equation}
    \max_{i\in C}\left[\max \left(|R_i-G_i|, |R_i-B_i|, |B_i-G_i|\right)\right] < \epsilon,
\end{equation}
where $C$ denotes pixels of a center crop of an image and $R_i$, $G_i$, $B_i$ denote the R, G, B channel values of pixel $i$ correspondingly. We perform center cropping in order to discard various external markings on the camera trap image (date, brand etc) that may be in color. We set $\epsilon = 10$. If low color variation is detected, we consider color related information in captions a hallucination. We use the following prompt to remove color information in this case:
\begin{shaded}
    \noindent \prompt[Prompt: ]{This is the description of an animal in a photograph: <LMM\_CAPTION>. Remove any mentions of color other than black or white. Answer:}
\end{shaded}

\textbf{Expert knowledge --} We enhance LMM captions with knowledge from our external knowledge base by adding details that may be visible in the image and removing information that directly contradicts the expert knowledge (extracted from Wikipedia articles). We use the following prompt for augmentation:
\newpage
\begin{shaded}
    \noindent \prompt[System message: ]{You are an AI assistant specialized in biology and providing \textcolor{red}{accurate and detailed descriptions of animal species.}}
    
    \prompt[Prompt: ]{This is an expert description of the appearance of an animal species: <EXPERT\_DESCR>. This is an image description of the same species I can see in a photograph: <LMM\_CAPTION>. Imagine that you can also see this photo and perform the following steps: \\
    1. Rewrite the image description by adding details from the expert description of the species that are visible in the photo. Make sure you only add details about body parts of the animal already present in the image description. \\
    2. Remove any information from the image description which directly contradicts the expert description. \\
    3. Do not mention the species name in the description and do not try to guess the species.\\
    Answer:}
\end{shaded}

\section{Instruction tuning data \label{apx:it}}
\begin{figure}
\centering
\begin{minipage}{.42\textwidth}
\centering
  \includegraphics[width=0.91\linewidth]{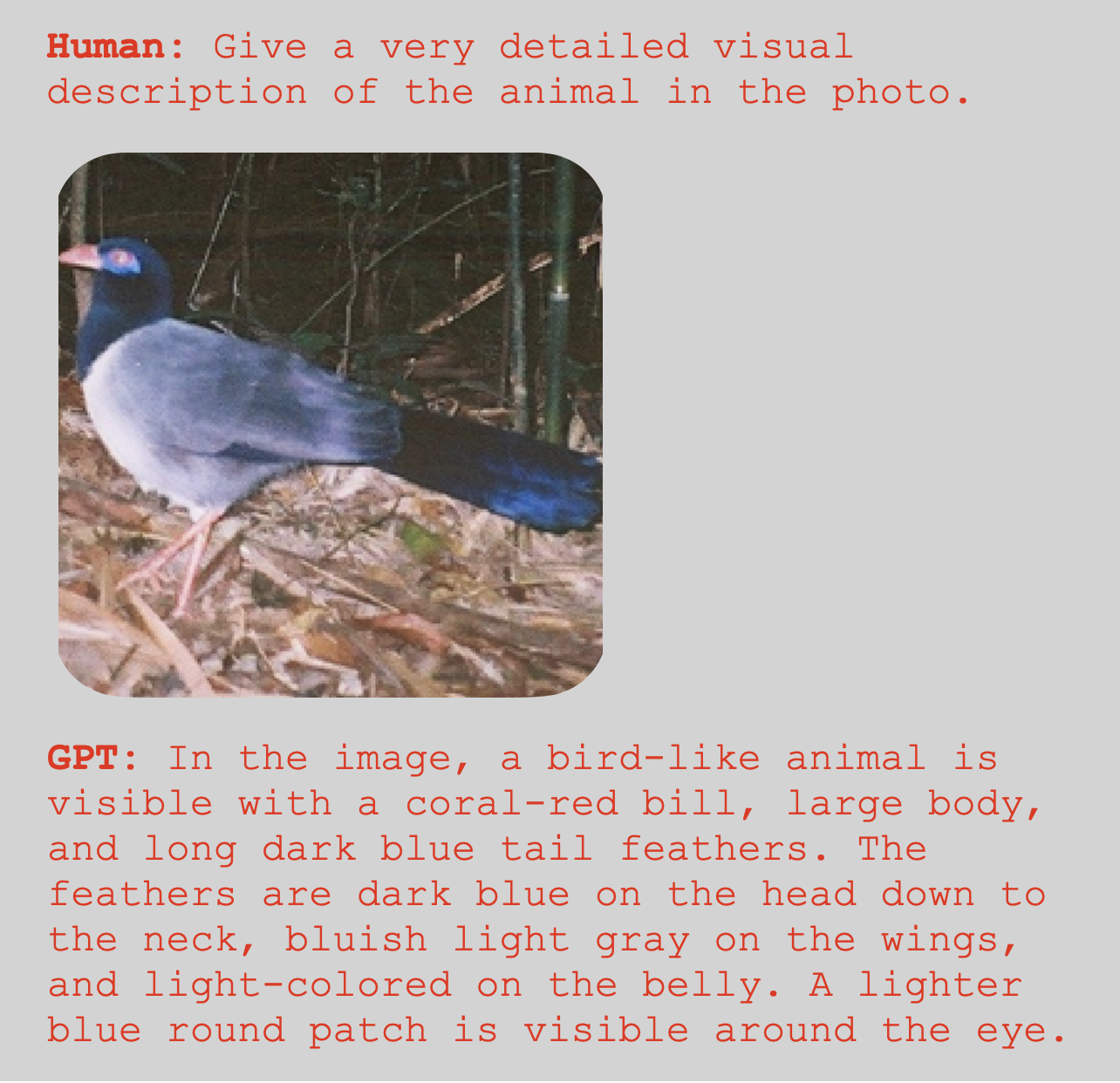}
  \captionof{figure}{A sample conversation from our instruction tuning pipeline.}
  \label{fig:it_sample}
\end{minipage}%
\quad
\begin{minipage}{.47\textwidth}
\centering
  \includegraphics[width=0.96\linewidth]{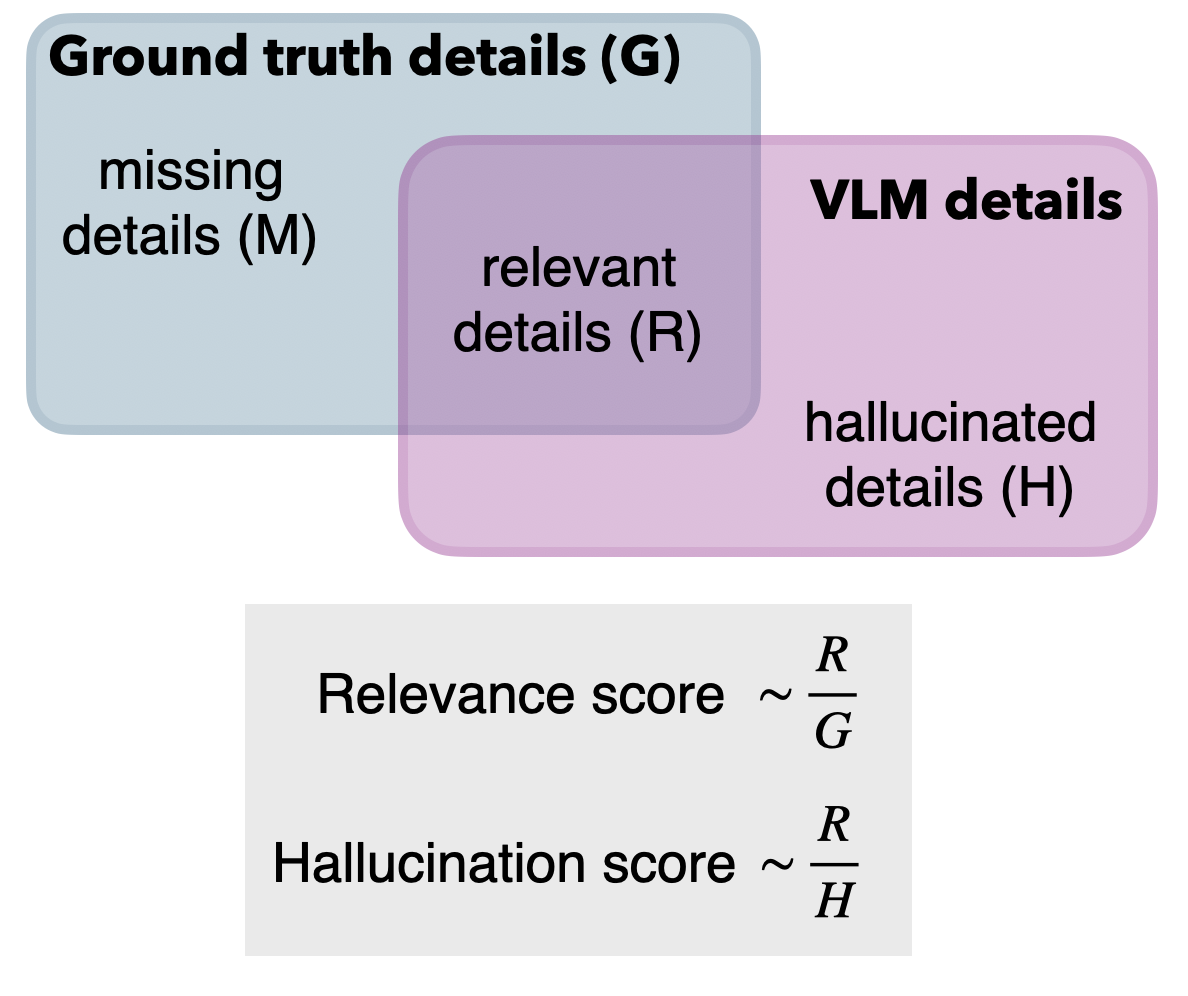}
  \captionof{figure}{Overview of GPT scores used to evaluate caption quality.}
  \label{fig:scoring}
\end{minipage}
\end{figure}

We build instruction tuning datasets of single-turn conversations where the model is asked to describe the animal in the image in great detail. In particular, we randomly sample from the following instructions:
\begin{shaded}
    \noindent \prompt{- Give a very detailed visual description of the animal in the photo.}

    \prompt{- Describe in detail the visible body parts of the animal in the photo.}

    \prompt{- What are the visual characteristics of the animal in the photo?}

    \prompt{- Describe the appearance of the animal in the photo.}

    \prompt{- What are the identifying characteristics of the animal visible in the photo?}

    \prompt{- How would you describe the animal in the photo?}

    \prompt{- What does the animal in the photo look like?}
\end{shaded}
We randomly place the image either before or after the instruction. The expected response is either the human-annotated caption we collected (\textit{Manual-1.5k}) or the automatically generated pseudo-caption corresponding to the image. A sample conversation is depicted in Figure \ref{fig:it_sample}.

\section{Self-consistency ablations}
\begin{figure}[t]
\centering
   \includegraphics[width=0.5\linewidth]{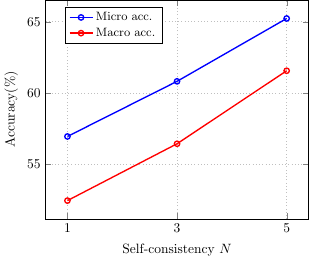}
  \caption{We observe consistent improvement as we increase the number of independent samples for inference (\datasetname{}, LLaVA-7B-P-gen1).\label{fig:sc_abl}}
\end{figure}

We analyze the effect of the number of independent LMM captions drawn ($N$) for inference in our pipeline. Due to the stochasticity in LMM-generated text, independently sampled captions may contain complementary information. To investigate this effect, we run experiments on the \datasetname{} with LLaVA-7B-P-gen1 and vary $N$ (Figure \ref{fig:sc_abl}). We observe steady increase in classification accuracy in both metrics as $N$ increases, hinting at potential for further zero-shot performance improvement with large $N$. The current pipeline leverages LLMs for description matching during inference which results in high compute costs with increasing $N$. We investigate opportunities to reduce the cost and increase $N$ in future work. 

\section{GPT-4 scoring\label{apx:scores}}
We perform qualitative analysis of generated captions via GPT-4 scoring. Assume that the image contains a set of ground truth details $G$ that can be found in an image description provided by an expert. Our LMM generates an image caption that covers some details from $G$ and we call this intersection \textit{relevant details} denoted by $R$. The rest of the details constitute the set of hallucinated details, denoted by $H$. An overview is depicted in Figure \ref{fig:scoring}. 

\textbf{Relevance score --} The relevance score evaluates the portion of ground truth details captured by the LMM caption. The more details are covered, the higher the score and no penalty is assigned for generating hallucinations. That is, the relevance score is proportional to $\frac{R}{G}$.The prompt used to evaluate the relevance score is as follows:
\begin{shaded}
\noindent \prompt[Prompt: ]{You are given two descriptions of an image: Description A and \textcolor{red}{Description B. Description A is the correct and accurate description of the image. Your job is to score on a scale from 1 to 10 how well Description B describes the image. Follow these rules: \\
1. Only give the maximum score of 10, if Description B contains all the information in Description A. \\
2. Only give the score of 1 if Description B contains no information that is given in Description A. \\
3. Otherwise, assign scores from 2 to 9 to assess how much information from Description A is mentioned in Description B (the higher score the more information from Description A is present in Description B). \\
4. Disregard any information in Description B that is not mentioned in Description A in your scoring. \\
5. Your answer is a single score from 1 to 10 without accompanying explanation of the score.
\\
Description A: <EXPERT\_DESCR>\\
Description B: <LMM\_CAPTION>\\
Your score:
}}
\end{shaded}

\textbf{Hallucination score --} This score measures how much additional information is included in the generated caption that is not present in the ground truth description and thus it is likely hallucinated or irrelevant. A higher score is assigned when the ratio $\frac{R}{H}$ is higher. That is as opposed to the relevance score, the hallucination score penalizes generated content not in the ground truth description. We use the following prompt to evaluate the hallucination score:
\begin{shaded}
\noindent \prompt[Prompt: ]{
You are given two descriptions of an image: Description A and Description B. Description A is the correct and accurate description of the image. Definition of a hallucination: a hallucination is a detail in Description B that is not mentioned in Description A. Your job is to score on a scale from 1 to 10 how accurately Description B describes the image, assigning higher score to descriptions with less hallucinations. Follow these rules:\\
1. Only give the maximum score of 10, if Description B contains all information from Description A and Description B does not contain any hallucinations.\\
2. Only give the score of 1 if Description B contains no information that is given in Description A, but may contain any number of hallucinated details.\\
3. Otherwise, assign scores from 2 to 9 to assess how much hallucinated information is present in Description B: the higher the score the less hallucinations are present in Description B.\\
5. Your answer is a single score from 1 to 10 without accompanying explanation of the score.\\
Description A: <EXPERT\_DESCR>\\
Description B: <LMM\_CAPTION>\\
Your score: 
}
\end{shaded}

\section{Human-annotated data collection \label{apx:manual}}
In order to measure the quality of pseudo-captions, we also collect a small dataset of human-captioned wildlife images. The dataset contains $2$ images of each species from the LILA camera traps and $1$ corresponding 'clean' image from the species' Wikipedia article. 

Instead of relying on expert annotators, we extract a list of visible features of each animal species in the dataset from Wikipedia using GPT-4. Then, the annotators are presented each of the features along with the image and are asked to select whether the feature is \textit{fully visible}, \textit{partially visible} or \textit{not visible} in the image. Furthermore, we ask annotators whether colors are discernible in the image in order to filter out color information from captions if needed.

Finally, we use GPT-4 to combine the features annotated as visible for each image into a descriptive image caption, and we apply post-processing to remove color information if it shouldn't be discernible.

\section{Description matching details \label{apx:match}}
In our proposed classification framework, we leverage an LLM to match the generated image description to an entry in the knowledge base. The label of the matched entry is used as the predicted label. In particular, we use GPT-4 to perform the matching step with the following prompt:
\begin{shaded}
\noindent \prompt[System message: ]{You are an AI expert in biology specialized in animal species identification. }\\
\prompt[Prompt: ]{<KNOWLEDGE\_BASE> \\
Question: You are given the following description of an animal: <LMM\_CAPTION>. What is the most likely animal being described from the following list: <SPECIES\_LIST>. Make sure your answer is a single word from the list <SPECIES\_LIST>. \\
Answer:
}
\end{shaded}
In the above prompt, the knowledge base is given as a list of entries in the form of \texttt{<SPECIES>:<DESCRIPTION>}.  We collect all \texttt{<SPECIES>} labels into a list \texttt{<SPECIES\_LIST>} in order to encourage the model to select one of the valid categories.

\textbf{Hierarchical predictions --} As described in Section \ref{sec:hierarch}, we use a hierarchical prediction scheme where we narrow down the potential number of fine-grained classes to at most $10$ before making a genus-level prediction.

\section{Calibration metrics\label{apx:calib}}
Assume a dataset with samples $(x, y)$ that are i.i.d. realizations of the random variables $X, Y \sim \mathbb{P}$. Further assume that a model predicts class label $y$ with probability $\hat{p}$. Then, the model is calibrated if $\mathbb{P}(Y=y|\hat{p}=p) = p$ for any $p \in [0,1]$ and label $y$. To measure calibration error, the probability interval $[0, 1]$ is first discretized into $n$ fixed bins denoted by $B_1, B_2, ..., B_n$. Let $acc(B_i)$ denote the ratio between correct predictions in bin $B_i$ and the total number of predictions that fall into $B_i$. Moreover, we define $conf(B_i)$ as the mean of probabilities in bin $B_i$. For a total number of $N$ samples in the test set, the following metrics are used to measure calibration error:
\begin{itemize}
    \item Expected Calibration Error (ECE): $$ECE = \sum_{i=1}^n \frac{|B_i|}{N} |acc(B_i) - conf(B_i)|$$
    \item Average Calibration Error (ACE): $$ACE = \sum_{i=1}^n \frac{1}{n} |acc(B_i) - conf(B_i)|$$    
    \item Maximum Calibration Error (MCE): $$MCE = \max_{i\in\{1..n\}} |acc(B_i) - conf(B_i)|$$
\end{itemize}